%% file: main.tex
\documentclass[sigconf]{acmart}

\usepackage{booktabs} %
\usepackage{multirow}
\usepackage{graphicx}
\usepackage{subfig}
\usepackage{placeins}
\usepackage{flushend} %
\usepackage[linesnumbered,ruled,vlined]{algorithm2e}
\usepackage{amsmath}

\DeclareMathOperator*{\argmax}{arg\,max}

\renewcommand{\vec}[1]{\mathbf{#1}}
\renewcommand{\matrix}[1]{\mathbf{#1}}

\newcommand{\pheadNoSpace}[1] {\noindent\textbf{#1.}} %
\newcommand{\pheadWithSpace}[1] {\vspace{1.25mm}\noindent\textbf{#1.}} %

\copyrightyear{2019}
\acmYear{2019} 
\setcopyright{iw3c2w3}
\acmConference[WWW '19]{Proceedings of the 2019 World Wide Web Conference}{May 13--17, 2019}{San Francisco, CA, USA}
\acmBooktitle{Proceedings of the 2019 World Wide Web Conference (WWW '19), May 13--17, 2019, San Francisco, CA, USA}
\acmPrice{}
\acmDOI{10.1145/3308558.3313656}
\acmISBN{978-1-4503-6674-8/19/05}

\usepackage{balance}

\begin{document}

\title[Be Concise and Precise]{Be Concise and Precise: Synthesizing~Open-Domain~Entity~Descriptions~from~Facts}

\author{Rajarshi Bhowmik}
\affiliation{
  \institution{Rutgers University - New Brunswick}
  \streetaddress{110 Frelinghuysen Road}
  \city{Piscataway}
  \state{New Jersey}
  \postcode{08854}}
\email{rajarshi.bhowmik@rutgers.edu}

\author{Gerard de Melo}
\affiliation{
  \institution{Rutgers University - New Brunswick}
  \streetaddress{110 Frelinghuysen Road}
  \city{Piscataway}
  \state{New Jersey}
  \postcode{08854}}
\email{gerard.demelo@cs.rutgers.edu}

\begin{abstract}
Despite being vast repositories of factual information, cross-domain knowledge graphs, such as Wikidata and the Google Knowledge Graph, only sparsely provide short synoptic descriptions for entities. Such descriptions
that briefly identify the most discernible features of an entity provide readers with a near-instantaneous understanding of what kind of entity they are being presented. They can also aid in tasks such as named entity disambiguation, ontological type determination, and answering entity queries. Given the rapidly increasing numbers of entities in knowledge graphs, a fully automated synthesis of succinct textual descriptions from underlying factual information is essential. To this end, we propose a novel fact-to-sequence encoder--decoder model with a suitable copy mechanism to generate concise and precise textual descriptions of entities. In an in-depth evaluation, we demonstrate that our method significantly outperforms state-of-the-art alternatives.
\end{abstract}

\ccsdesc[500]{Computing methodologies~Neural networks}
\ccsdesc[300]{Computing methodologies~Natural language generation}

\keywords{synoptic description generation; knowledge graphs; open-domain factual knowledge}

\maketitle
\input{body}
\bibliographystyle{ACM-Reference-Format}
\balance
\bibliography{bibliography}
\end{document}

%% file: body.tex
\section{Introduction}

\pheadNoSpace{Motivation}
A substantial percentage of online search requests involve named entities such as people, locations, businesses, etc. Search engines and digital personal assistants (e.g., Google Assistant, Alexa, Cortana, Siri) alike now extensively draw on \emph{knowledge graphs} as vast databases of entities and their properties. 

\begin{figure}[tb]
\centering
\includegraphics[width=0.9\linewidth]{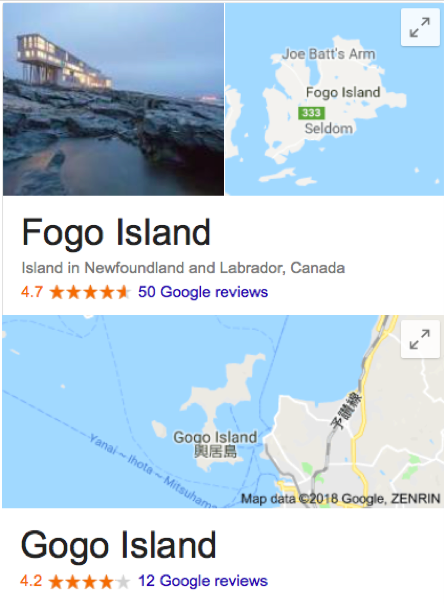}
\caption{An example of a missing description in Google's Knowledge Graph. Similar to \emph{Fogo Island}, a synoptic description for \emph{Gogo Island} could be \emph{Island in Ehime, Japan}.}
\label{fig:fogo_gogo}
\end{figure}

Most large cross-domain factual knowledge graphs, such as Wikidata \cite{Vrandecic:2014:WFC:Wikidata}, the Google Knowledge Graph, YAGO \cite{HoffartEtAlYAGO2}, and MENTA \cite{deMeloWeikum2010MENTA}, include short textual descriptions of entities. These can be provided in response to entity queries (such as \emph{Where is Fogo Island?}), e.g., as an informative direct answer or to enrich a knowledge panel given by a Web search engine such as Google, as shown in Figure~\ref{fig:fogo_gogo} for Fogo Island. Such textual descriptions effectively provide for a near-instantaneous human understanding of an entity. They can also be helpful in a number of linguistic tasks, including named entity disambiguation, and can serve as fine-grained ontological types in question answering and reasoning-driven applications.

Despite their eminent importance for information retrieval and other applications, these descriptions are only sparsely available, typically for the more well-known entities, leaving large numbers of entities with no such descriptions. For example, in Wikidata, which is used by Apple's Siri, as many as 7.2 million entities do not have any description in any language as of December 2018.\footnote{\url{https://tools.wmflabs.org/wikidata-todo/stats.php}}
Moreover, new entities are added frequently to the existing knowledge graphs. For example, the number of entities in Wikidata almost doubled in the last couple of years.

Given the large number of entities lacking textual descriptions, a promising solution is to fully automatically generate synoptic textual descriptions from the underlying factual data. Ideally, this automatic generation process should be able to synthesize very concise descriptions, retaining only a very small number of particularly discernible pieces of information about an entity. 
For example, in Wikidata and Google's Knowledge Graph, \emph{occupation} and \emph{nationality} are among the preferred attributes invoked in describing a person (e.g., \emph{Magnus Carlsen} as a \emph{Norwegian chess Grandmaster}). At the same time, however, an \emph{open-domain} solution is needed, in order to cope with the various different kinds of entities users may search for in a knowledge graph, and the various kinds of names that might appear in such descriptions. For instance, for Gogo Island in Figure~\ref{fig:fogo_gogo}, an appropriate description would perhaps refer to \emph{Ehime, Japan} as the prefecture in which the island is located. Moreover, apart from people and locations, users could likewise also search for paintings, films, or even asteroids.
Additionally, the generated descriptions should also be precise, coherent, and non-redundant.

\pheadWithSpace{Overview and Contributions}
In this paper, we present a novel fact-to-sequence encoder--decoder model
that solves this challenging task. Our deep neural model attempts to generate a textual description similar to those created in image and video captioning tasks \cite{LongChuangDeMelo2018CaptioningMultiFacetedAttention}, but instead starts from structured data. Our model is equipped with an explicit copy mechanism to copy fact-specific tokens directly to the output description. Since many names and words in the desired output descriptions are also present in the underlying facts, it is an intuitive design choice to selectively copy words from the facts to the description. The copy mechanism is important, as it is difficult to predict rare or unseen words based on mere statistical co-occurrences of words. In comparison to the  state-of-the-art, our model is more than 14x more parameter-efficient and gains 5--8 absolute percentage points across a number of metrics in an empirical evaluation.

This approach is substantially different from previous work on this topic. In contrast to some prior works that focus on generating Wikipedia-style summary descriptions from factual data \cite{LebretGA16, DBLP:journals/corr/AhnCPB16}, or textual verbalization of RDF triples \cite{Gardent-EtAl:2017:INLG2017, Ferreira2018}, our work is focused on synthesizing quickly graspable synoptic textual descriptions from factual knowledge. In most of the cases, these are single-phrase descriptions as compared to multi-sentence Wikipedia-style summaries.
Note that we restrict our work only to English description generation, as English is the most prevalent language in which descriptions are available in knowledge graphs.
Typically, the first few sentences of Wikipedia articles contain a detailed description of the entity, and one can attempt to generate brief textual descriptions by learning to reproduce the first few sentences based on the remaining article text. However, our goal is to make our method open-domain, i.e., applicable to any type of entity with factual information that may not have a corresponding Wikipedia-like article available. Indeed, Wikidata currently has more than 51 million items, whereas the English Wikipedia has only 5.6 million articles. 

Our specific contributions in this paper are as follows:
\begin{itemize}
    \item We introduce a fact-to-sequence encoder--decoder architecture that precisely copies fact-specific words using a two-step pointing and copying mechanism.
    \item Our model is 14x as parameter-efficient than the state-of-the-art models and 14--20x as parameter-efficient compared to competitive baselines.
    \item Through extensive evaluation, we show that our method outperforms other baselines with significant improvement, especially with regard to descriptions requiring open-domain name mentions.
\end{itemize}

\section{Model}
\begin{figure*}[tb]
\centering
\includegraphics[width=0.85\linewidth]{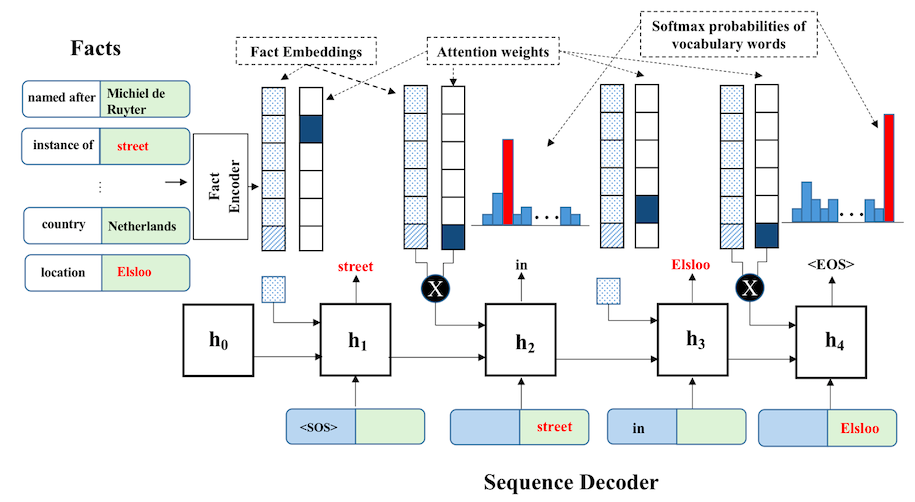}
\caption{Model architecture. For Wikidata item Q19345316 (\emph{Michiel de Ruyterstraat}), factual words such as \emph{street} and \emph{Elsloo} are directly copied from the underlying facts \emph{(Instance of, street)} and \emph{(location, Elsloo)}, respectively, while the general vocabulary words \emph{in} and \emph{<EOS>} are selected by a softmax classifier.}
\label{fig:model_architecture}
\end{figure*}

In order to enable concise and precise open-domain entity descriptions, 
we propose a generative model consisting of a fact-to-sequence encoder--decoder architecture, aided by a pointer network based copy mechanism. Our model comprises a positional fact encoder, a GRU-based sequence decoder, and a copy function to directly transfer factual words to the generated descriptions. Before delving into the details of these key components, however, we first introduce some essential terminologies that we shall use in the following.

\subsection{Preliminaries}

\pheadNoSpace{Property} Each subject entity $s \in \mathcal{E}$ has a number of properties $P \in \mathcal{P}$ associated with it. Each property name is described as a sequence of tokens $P = \{w_1^p, w_2^p, ..., w_{|P|}^p \}$. Note that these properties are predefined in a schema-based knowledge graph.

\pheadWithSpace{Property Value} Each property $P$ of entity $s$ has a corresponding value $O_P$, which is also a sequence of tokens $O_P = \{w_1^o, w_2^o, ..., w_{|O_P|}^o\}$. The property values could be another entity $e \in \mathcal{E}$, a date--time value, a string, or a numeric value, etc.

\pheadWithSpace{Fact} Each of the property--value pairs is deemed a \emph{fact}. Each subject entity $s \in \mathcal{E}$ has a number of different facts $\mathcal{F} = \{ f_1, f_2, \dots f_N\}$ that characterize it.

\pheadWithSpace{Description} A short textual description of entity $s$ is a sequence of tokens $\mathcal{D} = \{w_1^d, w_2^d, ..., w_{|\mathcal{D}|}^d\}$. Each word in the description $D$ can be a \emph{factual word}, or a \emph{vocabulary word}.

\pheadWithSpace{Factual Words} The factual words for each fact $f \in \mathcal{F}$, denoted as $\mathcal{V}_{f}$, are the words appearing in the property value.
Stop words (e.g., \emph{for}, \emph{of}, \emph{in} etc.) are excluded from the factual words vocabulary. Note that the sets of factual words are instance-specific, i.e., $\bigcup_f \mathcal{V}_{f}$ is different for different entities $s \in \mathcal{E}$.

\pheadWithSpace{Vocabulary Words} The vocabulary words, denoted as $\mathcal{V}$, could be a list of frequent words in the English dictionary. For our experiments, we rely on the 1,000 most frequent words appearing in our corpus as vocabulary words. Typically, the words that appear in the property names constitute this vocabulary. This is aligned with the expectation for a schema-based knowledge graph with fixed number of properties.  
Note that $\mathcal{V}_f \cap \mathcal{V} \neq \emptyset$, which indicates that there could be words in the property value that are also vocabulary words.

\subsection{Fact Encoder}
The fact encoder transforms a set of input facts $\mathcal{F} = \{ f_1, f_2, \dots f_N\}$ into \emph{fact embeddings} -- distributed vector representations of facts. For simplicity and efficiency, we use a positional encoding scheme proposed by Sukhbaatar et al.\ \cite{NIPS2015:Sukhbaatar} that was also adopted in other architectures \cite{Xiong:2016:DMN,BhowmikdeMelo2018}.

Using the positional encoding scheme, each fact $f_i$ is represented as $\vec{f}_i = \sum_j \vec{l}_j \odot \vec{w}_j^i$, where $\odot$ denotes element-wise multiplication, $\vec{l}_j$ is a column vector with the structure $l_{kj} =(1-\frac{j}{J})-(k/d)(1-2\frac{j}{J})$, with $J$ being the number of words in the fact, and $\vec{w}_j^i \in \mathrm{R}^d$ is the word embedding of the $j$-th word in fact $f_i$. If the word $w_i^j$ is not a vocabulary word, we replace it with the special token \emph{<UNK>}. This arrangement deals with the rare or unseen words that appear in the factual phrase $f_i$, which is a concatenation of words in the property name and property value.

We append the fact embeddings with a \emph{mean fact} -- a special fact that is responsible for generating vocabulary words. It is derived as the element-wise mean of the fact embeddings, i.e., %
$\vec{f}_{N+1} = \frac{1}{N} \sum \vec{f}_i$. We train the model to attend to the mean fact if it has to generate a vocabulary word in the output sequence. 

\subsection{Output Sequence Decoder}
At every time step $t$, the sequence decoder selects a fact. This selection governs the generation of the next word either from the corresponding factual words, or from the vocabulary words.

\subsubsection{Fact Selection}
Given the set of fact embeddings\\ $\{\vec{f}_1, \vec{f}_2, \dots , \vec{f}_{N+1}\}$ as input,
the decoder selects a fact at each time step $t$ using an attention mechanism. The attention scores are calculated as a likelihood distribution using the hidden state of the decoder in the previous time step and the fact embeddings. Formally,
\begin{equation}
    \vec{e}_i = \matrix{W}_2\,\mathrm{tanh}(\matrix{W}_1[\vec{f}_i; \vec{h}_{t-1}]) \textrm{ }\, \forall i \in \{1,N+1\},
\end{equation}
\begin{equation}
    P(f = f_i \mid \vec{f}_i, \vec{h}_{t-1}) = \frac{\mathrm{exp}(\vec{e}_i)}{\sum_{i' \in \{1, N+1\}} \mathrm{exp}(\vec{e}_{i'})},
\end{equation}
    
where $\vec{e}_i$ denotes the attention energy of the $i$-th fact, $[;]$ denotes the concatenation operation, and $\vec{W}_1 \in \mathrm{R}^{m \times 2d}$, $\vec{W}_2 \in \mathrm{R}^m$ are affine transformations of a 2-layer feed-forward network. 

We select the fact with maximum attention score at time step $t$, denoted as $f_t$ and its corresponding fact embedding $\vec{f}_t$ as
\begin{align}
    f_t &= \argmax_{i \in \{1, \dots, N+1\}} P(f = f_i \mid \vec{f}_i, \vec{h}_{t-1})\\
    \vec{f}_t &= \vec{f}_{f_t} 
\end{align}

\subsubsection{GRU Decoder}
We rely on Gated Recurrent Units (GRU), which is a Recurrent Neural Network (RNN) variant, as our preferred decoder. At each time step $t$, the decoder input is a concatenation of three vectors: the embedding $\vec{f}_t$ of a selected fact in the current time step, the embedding $\vec{w}_{t-1}$ of the vocabulary word at the previous time step, and a one-hot vector $\vec{v}_{t-1}$ corresponding to the position of the copied factual word in the previous time step. Note that since the generated word in the previous time step can either be a vocabulary word or a factual word, either $\vec{w}_{t-1}$ or $\vec{v}_{t-1}$ is set to a zero vector.

The input representation $\vec{x}_t = [\vec{f}_t; \vec{w}_{t-1}; \vec{v}_{t-1}]$ is fed to the GRU decoder to update the states.
\begin{equation}
    \vec{h}_t = \textrm{GRU}(\vec{x}_t, \vec{h}_{t-1})
\end{equation}

\subsubsection{Generating Vocabulary Words}
If the attention mechanism assigns the maximum score to the mean fact $f_{N+1}$, then the decoder generates a vocabulary word $w_t \in \mathcal{V}$. To generate the vocabulary word, we use the attention-weighted context $\vec{c}_t = \sum_i{\alpha_i\vec{f}_i}$
and the GRU output state $\vec{h}_t$. A concatenation of these two vectors are fed to a 2-layer feed-forward network with a non-linear ReLU activation applied to the hidden layer. Formally,
\begin{equation}
    \vec{o}_t = \matrix{W}_a\mathrm{ReLU}(\matrix{W}_b[\vec{c}_t; \vec{h}_t]),
\end{equation}
where $\matrix{W}_a$ and $\matrix{W}_b$ are affine transformations and $\mathrm{ReLU}(\vec{x}) = \max(\vec{0}, \vec{x})$.
Finally, a probability distribution over the vocabulary words is obtained by a softmax function and the word with the maximum probability is emitted by the decoder. Formally,
\begin{align}
    P(w \mid \vec{c}_t, \vec{h}_t) &= \mathrm{Softmax}(\vec{o}_t)\\
    w_t &= \argmax_{w \in \mathcal{V} }  P(w \mid \vec{c}_t, \vec{h}_t)
\end{align}

\subsubsection{Copying Factual Words}
The decoder copies factual words directly to the output when the fact selection process selects one of the $N$ facts. To facilitate the copy mechanism, the decoder must select the position index of the factual word within the factual phrase. The position index is predicted by a 2-layer feed-forward network that takes a concatenation of the selected fact embedding $\vec{f}_t$ and the output state of the GRU $\vec{h}_t$ as input. Then, the position index of the word to copy is determined as follows.
\begin{align}
    \vec{r}_t &= \matrix{W}_c\mathrm{ReLU}(\matrix{W}_d[\vec{f}_t; \vec{h}_t])\\
    P(n \mid \vec{f}_t, \vec{h}_t) &= \mathrm{Softmax}(\vec{r}_t)\\
    n_t &= \argmax_{n \in \{1, \dots, |\mathcal{V}_{f_t}|\}} P(n \mid \vec{f}_t, \vec{h}_t)\\
    w_t &= (\mathcal{V}_{f_t})_{n_t}
\end{align}
Here, $n_t$ is the position index of the factual word to copy and $\mathcal{V}_{f_t}$ is the sequence of factual words corresponding to fact $f_t$ and $(\mathcal{V}_{f_t})_{i}$ denotes the $i$-th item in that sequence.

\subsection{Training}
Our method requires selective copying of factual words to generate a description. In order to obviate the need for ground truth alignments of output description words with facts for training, we introduce an automated method of annotating each token in the description so as to align it to a matching fact. Specifically, we rely on a greedy string matching algorithm as detailed in Algorithm \ref{Algo:alignment} for this purpose. If a token is not aligned with any of the facts, it is annotated as a vocabulary word. However, if the token is neither present in the  vocabulary word set nor in the factual word set, it is assigned the special token \emph{<UNK>} to denote that it is neither a factual word nor a vocabulary word. The symbol $\textrm{NaF}$ in Algorithm \ref{Algo:alignment} indicates that such words are not aligned to any fact. Since in our implementation, we consider the mean fact as a source of vocabulary words, we align these words to the mean fact.

\begin{algorithm}[h]
\DontPrintSemicolon
\SetAlgoLined
\KwData{Input: $\mathcal{F}$, $\mathcal{D}$}
\KwResult{Ordered set of fact aligned description $\mathcal{\Tilde{D}}$ }
$\mathcal{\Tilde{D}} = \emptyset$\\
\For{$w \in \mathcal{D}$}{
    $\mathrm{factual\_flag} \gets \mathrm{False}$\\
    \For{$f \in \mathcal{F}$}{
        \If{$w \in \mathcal{V}_f$ $\mathrm{and}$ $\mathrm{factual\_flag = False}$}{
            $\mathcal{\Tilde{D}} \gets \mathcal{\Tilde{D}} \cup \{(w, f)\}$\\
            factual\_flag $\gets \mathrm{True}$\\
        }
    }
  \If{$\mathrm{factual\_flag = False}$}{
    \eIf{$w \in \mathcal{V}$}{
        $\mathcal{\Tilde{D}} \gets \mathcal{\Tilde{D}} \cup \{(w, \textrm{NaF})\}$\\
    }{
        $\mathcal{\Tilde{D}} \gets \mathcal{\Tilde{D}} \cup \{(\textrm{UNK}, \textrm{NaF})\}$\\
    }
   }
 }
 \caption{Algorithm for fact alignment of description}
 \label{Algo:alignment}
\end{algorithm}

Note that the greedy string matching algorithm used for fact alignment is a heuristic process that can be noisy. If a token in the output description appears in more than one fact, the string matching algorithm greedily aligns the token to the first fact it encounters, even if it is not the most relevant one. However, our manual inspection suggests that in most of the cases the alignment is relevant, justifying our choice of such a greedy approach.

Given the input facts $\mathcal{F}$ and the fact-aligned description $\tilde{\mathcal{D}}$, the model maximizes the log-likelihood of the observed words in the description with respect to the model parameters $\theta$,
\begin{equation}
    \theta^{*} = \argmax_\theta \log P(\mathcal{\tilde{D}} \mid \mathcal{F}),
\end{equation}
which can be further decomposed as
\begin{equation}
    \log P(\mathcal{\tilde{D}} \mid \mathcal{F}) = \sum_{t=1}^{|\mathcal{\tilde{D}}|} \log P(w_t \mid w_{1:t-1}, \mathcal{F}).
\end{equation}
Since the log-likelihood of the word $w_t$ also depends on the underlying fact selection, we can further decompose $P(w_t)$ as 
\begin{equation}
    P(w_t) = P(w_t \mid f_t, w_{1: t-1})\, P(f_t \mid w_{1:t-1})
\end{equation}
Therefore, we train our model end-to-end by optimizing the following objective function: 
\begin{equation}
    \mathcal{L}(\theta) = - \sum_{t=1}^{|\mathcal{D}|} \log P(w_t \mid f_t, w_{1:t-1}) - \sum_{t=1}^{|\mathcal{D}|} \log P(f_t \mid w_{1:t-1}).
\end{equation}
Note that the alignment $\mathcal{\tilde{D}}$ of the description $\mathcal{D}$ to the facts $\mathcal{F}$ provides a ground truth fact $f_t$ as each time step $t$ during training.

\section{Evaluation}
\begin{table*}[ht]
\centering
\caption{Experimental results for the WikiFacts10K-Imbalanced benchmark dataset}
\begin{tabular}{lccccccc}
\toprule
    Model & BLEU-1 & BLEU-2 & BLEU-3 & BLEU-4 & ROUGE-L & METEOR & CIDEr\\
\midrule
    Dynamic Memory-based Generative Model & 61.1 & 53.5 & 48.5 & 46.1 & 64.1 & 35.3 & 3.295\\ 
    Fact2Seq w. Attention & 62.8 & 56.3 & 53.0 & 52.7 & 63.0 & 35.0 & 3.321\\
    NKLM & 34.7 & 27.7 & 29.1 & 29.0 & 44.1 & 20.1  & 1.949\\
    \midrule
    Our Model & \textbf{61.9} & \textbf{57.3} & \textbf{55.6} & \textbf{54.3} & \textbf{64.9} & \textbf{36.3} & \textbf{3.420}\\
    -- without Positional Encoder & 58.9 & 51.9 & 46.3 & 42.5 & 63.4 & 33.7 & 3.126\\
    -- without Mean Fact & 58.0 & 52.2 & 49.5 & 48.6 & 64.3 & 34.8 & 3.139 \\
    -- Copy Only & 26.8 & 21.2 & 16.5 & 11.8 & 45.3 & 20.6 & 1.766\\
    
\bottomrule
\end{tabular}
\label{tab:WikiFacts10k-Imbalanced}
\end{table*}
\begin{table*}[ht]
\centering
\caption{Experimental results on the WikiFacts10K-OpenDomain benchmark dataset }
\begin{tabular}{lccccccc}
\toprule
    Model & BLEU-1 & BLEU-2 & BLEU-3 & BLEU-4 & ROUGE-L & METEOR & CIDEr\\
\midrule
    Dynamic Memory-based Generative Model & 57.8 & 50.7 & 43.6 & 39.7 & 67.6 & 34.9 & 3.556\\ 
    Fact2Seq w. Attention & 62.7 & 56.3 & 50.0 & 46.2 & 67.7 & 35.8 & 3.629\\
    NKLM & 47.9 & 42.1 & 37.5 & 32.3 & 57.4 & 25.9 & 2.958\\
    \midrule
    Our Model & \textbf{68.2} & \textbf{61.8} & \textbf{56.6} & \textbf{51.9} & \textbf{70.0} & \textbf{37.3} & \textbf{4.084}\\
    -- without Positional Encoder & 
    66.2 & 58.4 & 51.8 & 45.9 & 67.9 & 34.7 & 3.717 \\
    -- without Mean Fact & 62.4 & 58.0 & 55.1 & 51.8 & 67.7 & 36.0 & 3.856\\
    -- Copy Only & 37.5 & 26.4 & 18.4 & 10.9 & 55.2 & 24.8 & 2.136\\
\bottomrule
\end{tabular}
\label{tab:WikiFacts10k-OpenDomain}
\end{table*}

\subsection{Baselines}
We compare the performance of our model against a number of competitive baselines, including recent state-of-the-art deep neural methods.

\subsubsection{Dynamic Memory-based Generative Model.} We use the model proposed by Bhowmik and de Melo \cite{BhowmikdeMelo2018} as one of our baseline methods, because that model claims to solve essentially the same problem using a dynamic memory-based generative model, achieving state-of-the-art results. The dynamic memory component memorizes  how much information from a particular fact is used by the previous memory state and how much information of a particular fact is invoked in the current context of the output sequence. However, through our experiments and analysis, we show that this approach is unable to obtain precise open-domain descriptions with arbitrary tokens. 
We use the original implementation\footnote{\url{https://github.com/kingsaint/Open-vocabulary-entity-type-description}}.

\subsubsection{Fact-to-Sequence with Attention.} The comparison with this baseline method can be deemed as an ablation study, in which the decoder has no access to the copy mechanism. This baseline resembles the standard sequence-to-sequence with attention mechanism proposed by Bahdanau et al.\ \cite{DBLP:journals/corr/BahdanauCB14}. However, unlike in their work, the input to this model consists of positionally encoded discrete facts.

Each word in the output sequence is predicted by providing the attention-weighted fact embeddings and the previous hidden state as an input to the GRU decoder. Then, the output of the decoder is concatenated with the attention-weighted fact embeddings and passed through a 2-layer feed-forward network with a ReLU activation. Finally, the output of the MLP is fed to a softmax classifier that outputs a probability distribution for the combined vocabulary of factual and non-factual words $\mathcal{V} \cup \{\bigcup_{s \in \mathcal{E}} \mathcal{V}_f^s\}$. The word with the maximum probability is emitted by the decoder. 

Note that Bhowmik and de Melo \cite{BhowmikdeMelo2018} also implements a ``Fact2Seq w.\ Attention'' baseline. However, one key difference is that the baseline considered here has a skip-connection from the attention-weighted fact embeddings, concatenating them with the output of the GRU decoder that is then passed through a 2-layer feed-forward network. The experimental results suggest that our variant with this skip-connection leads to substantially more accurate descriptions.

\subsubsection{Neural Knowledge Language Model (NKLM)} We compare against Ahn et al.'s Neural Knowledge Language Model \cite{DBLP:journals/corr/AhnCPB16}, which is able to generate Wikipedia-style multi-sentence summary paragraphs for movie actors. Although the descriptions in our case are much shorter, we adopted their model as a representative baseline for methods yielding multi-sentence summaries, as the tasks are similar in nature. NKLM also adopts a copy mechanism but the decision about whether to copy is made by a binary gating variable that is provided as an additional label by augmenting the dataset during training. By predicting whether the word to generate has an underlying fact or not, the model can generate such words by copying from the selected fact.
On the contrary, our model decides to copy whenever a fact other than the mean fact is selected. We implemented and trained their model with the benchmark datasets. Their model also requires an alignment of descriptions to the underlying facts. Additionally, NKLM relies on pre-trained TransE \cite{NIPS2013:Bordes} embeddings of objects and relations to obtain the fact embeddings that are provided as inputs to the model. 

\subsection{Dataset}
\begin{table}[ht]
    \centering
    \begin{tabular}{lr|lr}
        \hline
        \multicolumn{2}{c|}{WikiFacts10k-OpenDomain} & \multicolumn{2}{c}{WikiFacts10k-Imbalanced}\\
        \hline
        Instance of & Pct. & Instance of & Pct. \\
        \hline
        human & 11.45\% & human & 69.17\% \\
        painting & 8.73\% & painting  & 2.02\%\\
        commune of france & 6.41\% & commune of france & 1.43\%\\
        film & 6.27\% & scientific article & 1.36\%\\
        scientific article & 5.56\% & film & 1.34\%\\
        encyclopedic article & 3.30\% & encyclopedic article & 0.89\% \\
        asteroid & 2.89\% & asteroid & 0.62\%\\
        taxon & 2.58\% & taxon & 0.57\%\\
        album & 2.19\% & road & 0.52\% \\
        road & 2.02\% & album & 0.46\% \\
        \hline
    \end{tabular}
    \caption{Frequency distribution of the top-10 domains in the two datasets.}
    \label{tab:domain_frequency}
\end{table}

For our experiments, we rely on two benchmark datasets, each of which consists of 10K entities with at least 5 facts and an English description.
The first of these datasets, denoted as \emph{WikiData10K-Imbalanced}, is the one used in Bhowmik \& de Melo (2018) \cite{BhowmikdeMelo2018}.
However, since the entities in that dataset were randomly sampled from the Wikidata RDF dump \cite{E+14:WikidataRDF}, the ontological types of the sampled entities have a long-tail distribution, while an overwhelming 69.17\% of entities are instances of \emph{human}. This highly skewed distribution makes the dataset biased towards a particular domain. To decrease the percentage of such entities in the dataset, we created a new dataset, \emph{WikiData10K-OpenDomain}, in which the instances of \emph{human} are downsampled to 11.45\% to accommodate more instances of other ontological types as evinced by the frequency distribution in Table \ref{tab:domain_frequency}. These datasets, both available online\footnote{\url{https://github.com/kingsaint/Wikidata-Descriptions}}, are split into training, dev., and test sets in a 80:10:10 ratio.

\subsection{Metrics}

Following previous work, we use the automatic evaluation metrics BLEU (1 - 4) \cite{ACL2002:BLEU}, ROUGE-L \cite{Lin:2004}, METEOR \cite{Lavie:2007:MAM:1626355.1626389}, and CIDEr \cite{DBLP:journals/corr/VedantamZP14a} for a quantitative evaluation of the generated descriptions with respect to the ground truth descriptions provided in the benchmark data. These metrics are widely used in the literature for the evaluation of machine translation, text summarization, and image captioning. BLEU is a precision-focused metric, ROUGE-L is a recall-based metric, METEOR uses both precision and recall with more weight on recall than precision, and CIDEr considers TF-IDF-weighted n-gram similarities.
Following standard practice, as in the official implementation\footnote{\url{https://github.com/vrama91/cider}}, the raw scores for CIDEr are multiplied by a factor of 10 for better readability, yielding values in $[0,10]$. For the same reason, following standard practice, the scores for other metrics are multiplied by 100, yielding scores in the range of $[0,100]$.
Note that typically these metrics rely on more than one human-written ground truth output per instance for evaluation. The lack of any alternative descriptions in the benchmark datasets implies that the generated descriptions are each evaluated on a single ground truth description. Additionally, these metrics take a very conservative approach in that they look for overlapping words, word alignment, longest common subsequence etc.

\subsection{Experimental Setup}
Our model and all other baselines are trained for a maximum 25 epochs. We report the results of the best performing models on the dev set. For NKLM and the dynamic memory-based generative model, we use the default hyper-parameter settings used by the authors. For NKLM, following the authors, we obtain the fact embeddings by concatenating the embeddings of object entity and relation that are obtained by using the TransE embedding model. For our model and the fact-to-sequence with attention baseline, we fix the embedding dimensions of facts and words to 100. The hidden layers of the GRU and the 2-layer feed-forward networks are 100-dimensional. We use Adam as our optimizer with a fixed learning rate of 0.001. We fix the maximum number of facts to 60 including the mean fact. For instances with less that 60 facts, we resort to a suitable masking to obtain the attention scores of the relevant facts. The maximum number of factual words for each fact is limited to 60.

We use the 1,000 most frequent words in the dataset as our default vocabulary. One can also consider using a much larger vocabulary of frequent English words. However, for our experiments we found it unnecessary. Still, because of this restricted vocabulary, our model may occasionally sparsely generate <UNK> tokens, which we remove from the generated description.

The datasets and the PyTorch implementations of all our experiments are available online.\footnote{\url{https://github.com/kingsaint/Wikidata-Descriptions}}

\subsection{Results}
Tables \ref{tab:WikiFacts10k-Imbalanced} and \ref{tab:WikiFacts10k-OpenDomain} provide the evaluation results of our model and the baselines on the WikiFacts10k-Imbalanced and WikiFacts10k-OpenDomain datasets, respectively. Our model outperforms the state-of-the-art dynamic memory-based generative model by more than 8 BLEU-4 points on the imbalanced dataset. We also observe 1 to 7 point gains on the scores across all other metrics.
Similar trends are observed for the more challenging WikiFacts10k-OpenDomain dataset, in which our model improves upon the dynamic memory-based generative model by 12.2 points in terms of the BLEU-4 metric. For this dataset, we also observe 5 to 8 point gains across all other metrics. A substantial gain in the BLEU scores indicates that our model can generate longer sequences more accurately than the state-of-the-art model.

Furthermore, we observe that our attention-based fact-to-sequence model also outperforms the dynamic memory-based generative model in most of the metrics across both the datasets. This observation shows that one can do without a dynamic memory module for memorizing fact specific information. It turns out that a GRU-based architecture with an intelligently chosen encoding of the input and an extra skip-connection in the decoder has sufficient memory to retain fact-specific information required to generate words in the output.

The NKLM model by Ahn et al.\ obtains sub-par results on both datasets, particularly the WikiFacts10k-Imbalanced data, although the original paper focused on descriptions of humans. This shows that their method for generating a short text is unsuitable for our demanding task of generating very concise entity descriptions. Additionally, the pre-trained TransE embeddings required to obtain fact embeddings are not available for previously unseen entities in the test set, thus severely restricting the model's ability to cope with unknown entities.
\subsection{Ablation Study} In addition to comparing our method against a number of competitive baselines, we perform a series of ablation studies to evaluate the effect of different components within our model. 
The results of these experiments are included in Tables \ref{tab:WikiFacts10k-Imbalanced} and \ref{tab:WikiFacts10k-OpenDomain}.

\subsubsection{Without Positional Encoder} To understand the effect of the positional encoder, we replace the positional encoder with the average pooling of the constituent word embeddings of the facts to obtain the fact embeddings.
The differences between the BLEU scores of our model and this ablated version indicates that the position of the words in the factual phrases indeed plays an important role in encoding a fact that provides a useful signal to the decoder.

\subsubsection{Without Mean Fact} To understand the effect of mean fact in our model, we replace the mean fact with a fixed embedding vector of dimension $d$. The elements of this vector are sampled from a uniform distribution $\mathcal{U}(-\frac{1}{\sqrt{d}}, \frac{1}{\sqrt{d}})$ and do not change during training. 
The result of this experiment suggests that the role of the mean fact in our model is not of just that of a \emph{sentinel} responsible for generating general vocabulary words. Rather, it encodes some useful semantic information that the model can exploit. 
\subsubsection{Copy Only} This experiment is an ablation study that we perform to understand how precisely a model can generate descriptions if it only copies factual words. This baseline is a restricted version of our model, in which the model never generates any general vocabulary word. 
The much inferior performance of this model shows that merely copying factual words to the output does not yield precise descriptions. This further demonstrates the importance of a model that dynamically decides whether to copy words from the input facts or instead emit general vocabulary words.
\subsection{Parameter Efficiency}
\begin{table}[htbp]
    \centering
    \begin{tabular}{lr}
        \toprule
        Model & \#Parameters \\
        \midrule
        Dynamic memory-based generative model & 14,197,741\\
        Fact2Seq w. Attention & 14,159,561\\
        Neural Knowledge Language Model (NKLM) & 20,569,361\\
        \midrule
        Our Model & 979,986\\
        \bottomrule
    \end{tabular}
    \caption{Comparison of the number of learnable parameters.}
    \label{tab:parameters}
\end{table}

Table \ref{tab:parameters} shows the number of learnable parameters for our model as well as the baselines. Our model is 14x more parameter-efficient than the competitive baselines. As there are fewer parameters to learn, this drastically improves the average training time of the model as compared to the other baselines. The number of parameters depends on the vocabulary size of the output softmax layer and the input to the word embeddings. Reducing the size of the softmax vocabulary to the most frequent words and making the model copy fact-specific words directly from the input contributes to the parameter efficiency and faster training of the model. 

\subsection{Importance of Fact Alignment of Descriptions} 
\begin{figure*}[htbp]
    \centering
    \vspace*{2cm}
    \subfloat[Our Model]{{\includegraphics[width=0.47\textwidth]{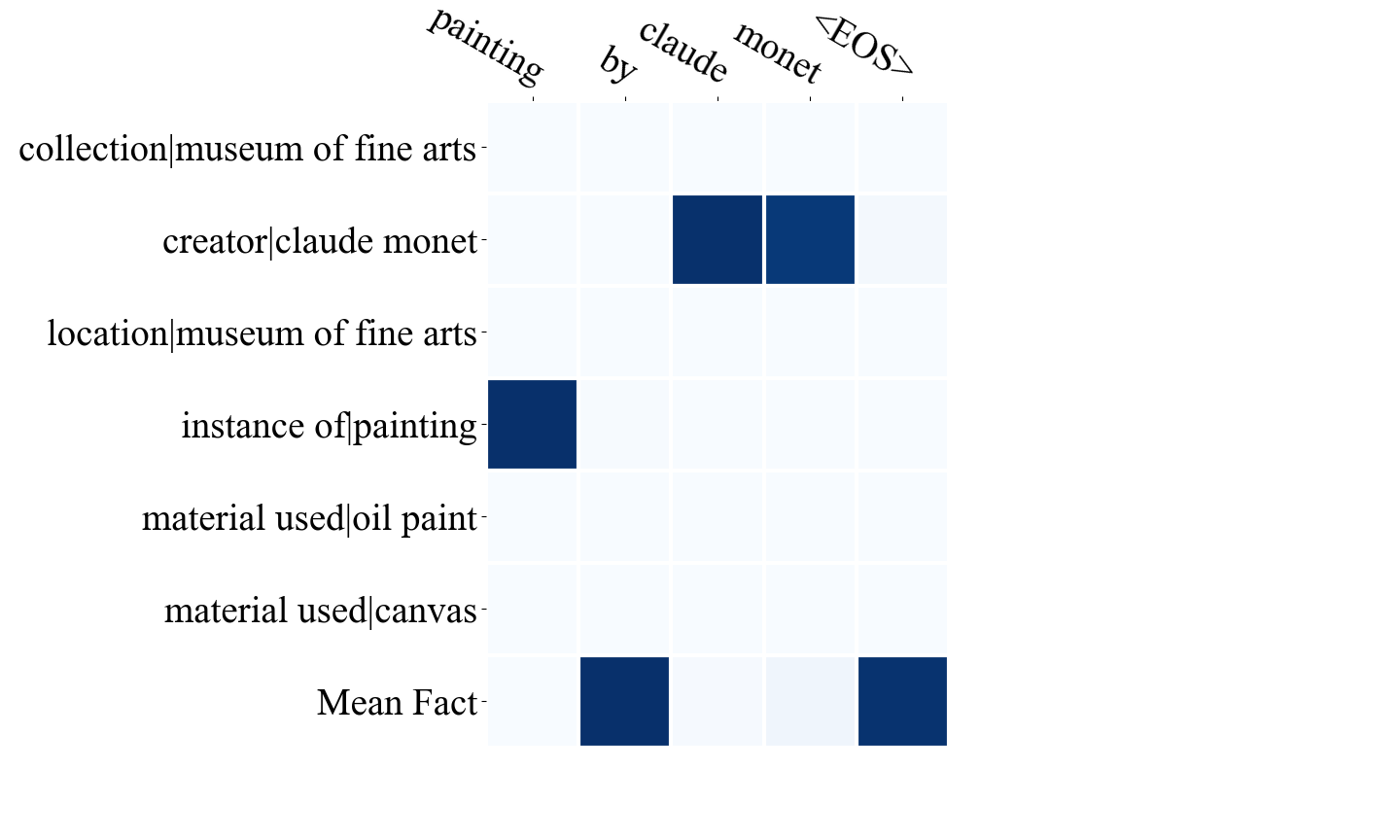} }}%
    \qquad
    \subfloat[NKLM]{{\includegraphics[width=0.47\textwidth]{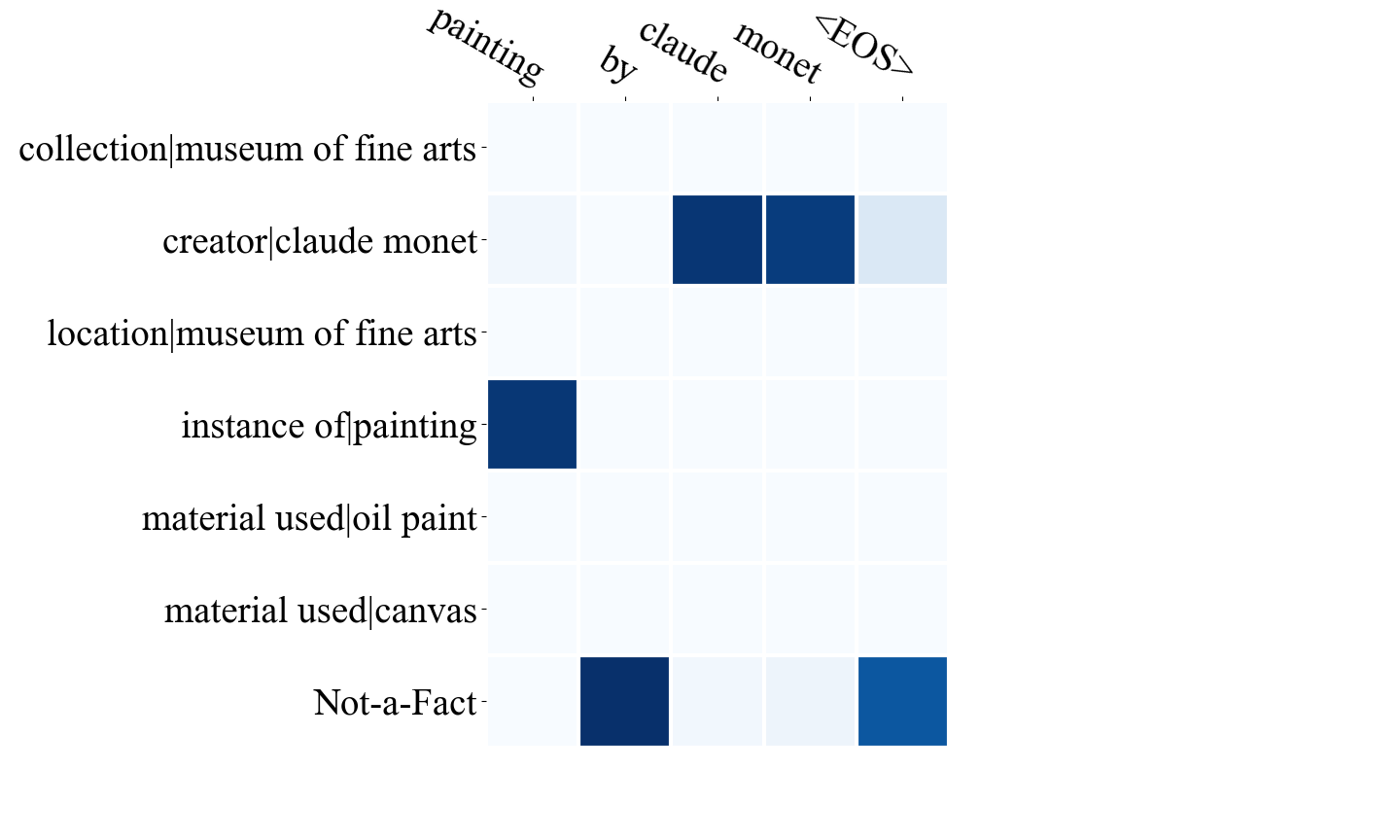} }}%
    \qquad
    \subfloat[Fact2Seq]{{\includegraphics[width=0.47\textwidth]{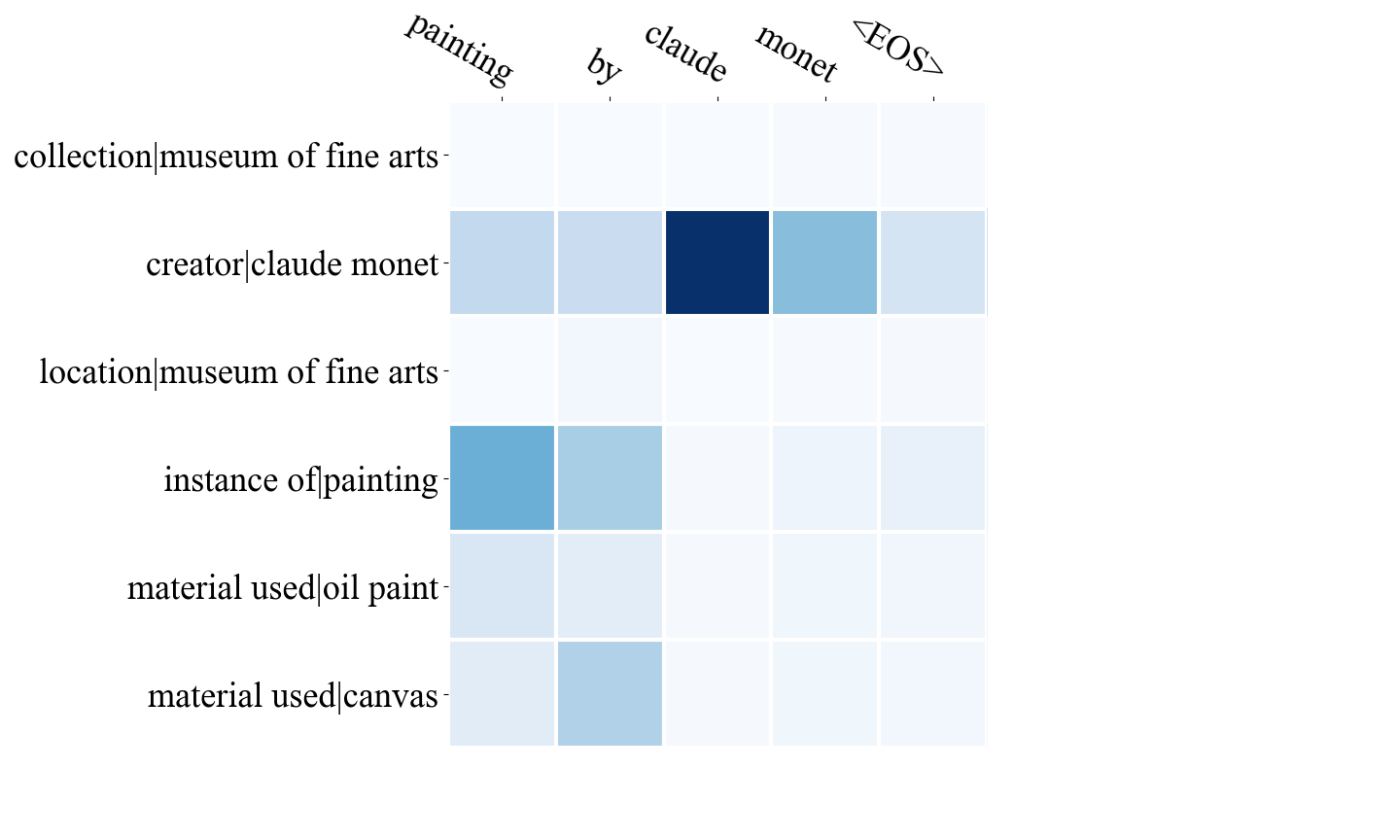} }}%
    \qquad
    \subfloat[Dynamic Memory-based Generative Model]{{\includegraphics[width=0.47\textwidth]{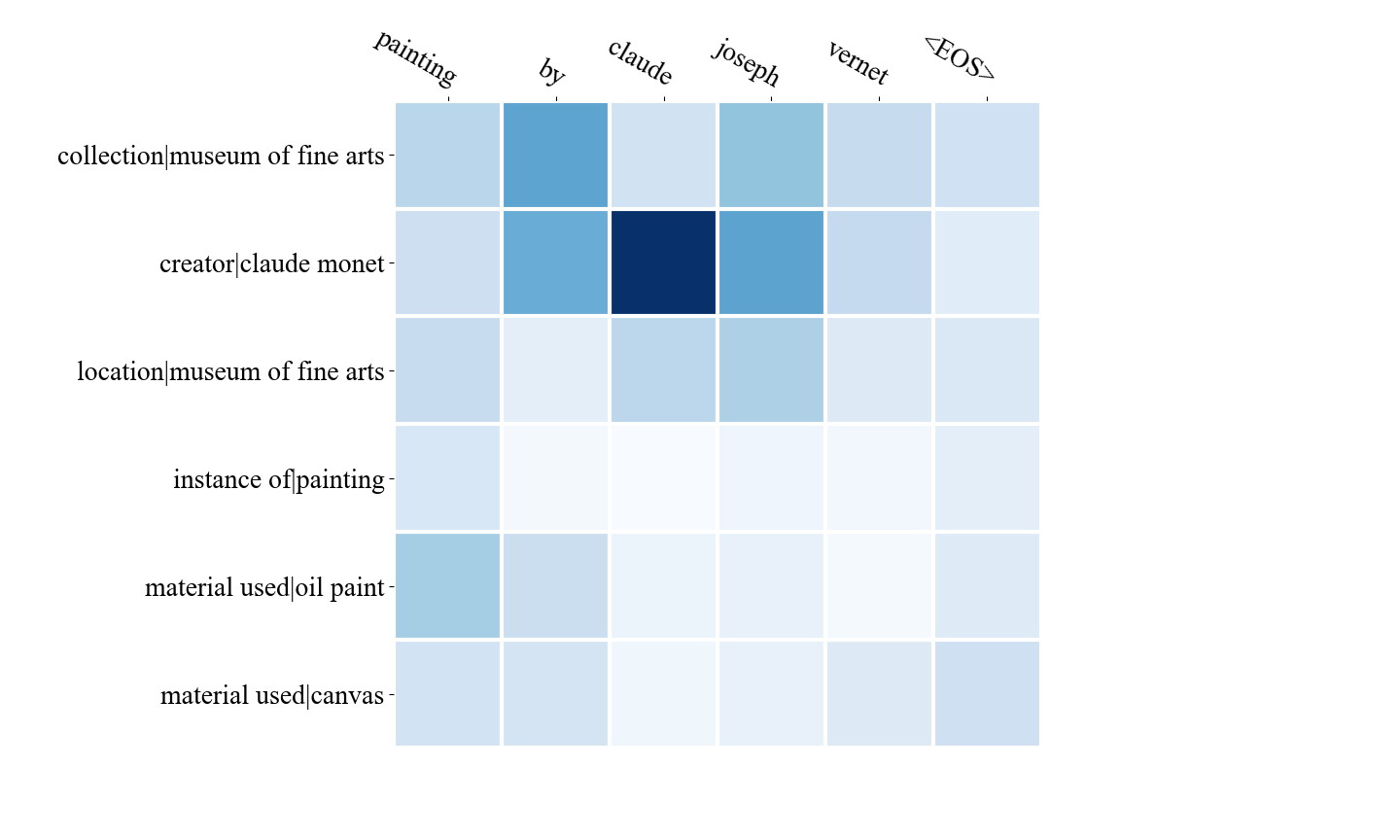}}}%
    \caption{A visualization of attention weights for selecting relevant facts.}%
    \vspace*{2cm}
    \label{fig:heatmap}%
\end{figure*}

The alignment of facts to the description teaches the model to choose the right fact for generating a factual word. As shown in Fig.~\ref{fig:heatmap}, for each word of the generated description, our model precisely selects the relevant underlying fact. The NKLM also shows a similar effect. However, the dynamic memory-based generative~model does not always pick the most relevant fact, although it might occasionally generate the right word  due to the statistical co-occurrence of the words in the training set.

\begin{table*}[htbp]
    \centering
    \begin{tabular}{lccccccc}
        \toprule
    Model & BLEU-1 & BLEU-2 & BLEU-3 & BLEU-4 & ROUGE-L & METEOR & CIDEr\\
\midrule
    Dynamic Memory-based Generative Model & 39.5 & 31.9 & 22.3 & 14.8 & 55.9 & 23.6 & 2.319\\ 
    Fact2Seq w. Attention & 52.8 & 44.5 & 33.5 & 24.0 & 56.9 & 24.1 & 2.337\\
    Neural Knowledge Language Model & 74.3 & 68.3 & \textbf{63.1} & \textbf{57.2} & 73.9 & 40.1 & 5.348\\
    \midrule
    Our Model & \textbf{75.4} & \textbf{69.1} & 62.8 & 55.5 & \textbf{76.8} & \textbf{41.4} & \textbf{5.387}\\
    - Copy Only & 42.7 & 26.3 & 15.4 & 0.0 & 57.9 & 27.1 & 2.089\\
\bottomrule
    \end{tabular}
    \vspace*{3pt}
    \caption{Experimental results for the subset of ontological types that require explicit copying of factual words.}
    \label{tab:subset_result}
\end{table*}
\begin{table*}[htpb]
    \centering
    \begin{tabular}{|c|c|l|l|l|}
        \hline
        Instance Of & Ground Truth Description & \multicolumn{2}{c}{Generated Description}\\
        \hline
        \multirow{8}{*}{Album} & \multirow{4}{*}{album by hypocrisy} & Dynamic Memory & album by michelangelo\\ 
        & & Fact2Seq & album by song\\
        & & NKLM & album by hypocrisy hypocrisy\\
        & & Our Model & album by hypocrisy\\
        \cline{2-4}
        & \multirow{4}{*}{album by canadian country
        music group family brown} & Dynamic Memory & czech hong kong by rapper laurana \\
        & & Fact2Seq & sparta album \\
        & & NKLM & album by family brown\\
        & & Our Model & album by family brown\\
        \hline
        \multirow{4}{*}{Book} & \multirow{4}{*}{science fiction novel by richard k morgan} & Dynamic Memory & science fiction novel by\\ 
        & & Fact2Seq & episode of a book by bernard dyer\\
        & & NKLM & fiction by richard k\\
        & & Our Model & novel by richard k\\
        \hline
        \multirow{8}{*}{Painting} & \multirow{4}{*}{painting by hendrick cornelisz van vliet} & Dynamic Memory & painting by cornelis de vos\\
        & & Fact2Seq & painting by abraham van (ii) switzerland\\
        & & NKLM & painting by hendrick cornelisz\\
        & & Our Model & painting by hendrick cornelisz van vliet\\
        \cline{2-4}
        & \multirow{4}{*}{painting by eustache le sueur} & Dynamic Memory & painting by thomas hovenden\\
        & & Fact2Seq & painting by california\\
        & & NKLM & painting by eustache le\\
        & & Our Model & painting by le sueur sueur\\
        \hline
        \multirow{8}{*}{Road} & \multirow{4}{*}{highway in new york} & Dynamic Memory & highway in new york\\ 
        & & Fact2Seq & highway in new york\\
        & & NKLM & <UNK> highways york\\
        & & Our Model & highway in new york\\
        \cline{2-4}
        & \multirow{4}{*}{road in england} & Dynamic Memory & area in the london borough of croydon\\
        & & Fact2Seq & of in london\\
        & & NKLM & road road in the church of england\\
        & & Our Model & road in the area london \\
        \hline
        \multirow{8}{*}{Sculpture} & \multirow{4}{*}{sculpture by antoine coysevox} & Dynamic Memory & sculpture by frederick william pomeroy\\ 
        & & Fact2Seq & sculpture by unknown singer\\
        & & NKLM & sculpture by antoine coysevox\\
        & & Our Model & artwork by antoine coysevox\\
        \cline{2-4}
        & \multirow{4}{*}{sculpture by donatello} & Dynamic Memory & by henry and final by the carducci\\ 
        & & Fact2Seq & sculpture by statue kreis comedy\\
        & & NKLM & sculpture by donatello donatello nilo\\
        & & Our Model & sculpture by donatello\\
        \hline
        \multirow{8}{*}{Single/ Song} & \multirow{4}{*}{1967 gilbert b\char"00E9caud song} & Dynamic Memory & 2014 by 2012 of the czech 2014 by czech\\ 
        & & Fact2Seq & song\\
        & & NKLM & song by gilbert b\char"00E9caud\\
        & & Our Model & song by gilbert b\char"00E9caud\\
        \cline{2-4}
        & \multirow{4}{*}{single} & Dynamic Memory & song by dutch by 2014 municipality\\ 
        & & Fact2Seq & 1980 song by northern song\\
        & & NKLM & single by michael jackson\\
        & & Our Model & song by michael jackson\\
        \hline
        \multirow{8}{*}{Street} & \multirow{4}{*}{street in boelenslaan} & Dynamic Memory & street in richard\\ 
        & & Fact2Seq &street in collection\\
        & & NKLM & street in boelenslaan\\
        & & Our Model & street in achtkarspelen\\
        \cline{2-4}
        & \multirow{4}{*}{street in echt} & Dynamic Memory & street in singer\\ 
        & & Fact2Seq & street in one\\
        & & NKLM & street in echt\\
        & & Our Model & street in echt\\
        \hline
    \end{tabular}
    \caption{Examples from the subset of ontological types that benefits from copy mechanism.}
    \label{tab:example}
\end{table*}
\begin{table}[htpb]
    \centering
    \begin{tabular}{lll}
        \toprule
        Item & Instance of & Generated Description \\
        \midrule
        Q11584386 & Human & japanese tarento\\
        Q2198428 & Human & netherlands businessperson\\
        Q3260917 & Human & french military personnel\\
        \midrule
        Q1494733 & Painting & painting by august macke\\
        Q16054316 & Painting & painting by liselotte \\
        & & schramm-heckmann\\
        Q15880468 & Painting & painting by emile wauters\\
        \midrule
        Q10288648 & Book & book by izomar camargo guilherme\\
        Q10270545 & Book & novel by antonin kratochvil\\
        Q10272202 & Book & novel by jose louzeiro\\
        \midrule
        Q1001786 & Street & street in budapest\\
        Q10552208 & Street & street in orebro\\
        Q10570752 & Street & street in malmo municipality\\
        \bottomrule
    \end{tabular}
    \caption{Examples of generated descriptions for the Wikidata entities with missing descriptions.}
    \label{tab:missing_desc_example}
\end{table}

\subsection{Significance of the Copy Mechanism} 
The copy mechanism enables our model to copy fact-specific rare or previously unseen words directly to the output sequence, generating factually correct descriptions. To demonstrate the positive effect of this copy mechanism, we select instances from a subset of 8 ontological types that tend to require explicit copying of factual words. This subset accounts for 19.6\% of the test set. Table \ref{tab:example} shows some examples from this subset. We also perform an automatic evaluation on this subset and provide the results in Table \ref{tab:subset_result}. Both the models with copy mechanism significantly outperform the baselines lacking any explicit copy mechanism. These results demonstrate that a suitable copy mechanism generates far more precise open-domain descriptions.

\section{Related Work}
In the following, we review previous research and describe how it differs from the task and approach we consider in this paper.

\subsection{Text Generation from Structured Data}
Lebret et al.\ \cite{LebretGA16} take Wikipedia infobox data as input and train a neural language model that, conditioned on occurrences of words in the input table, generates biographical sentences as output.
In a similar vein, Ahn et al.\ \cite{DBLP:journals/corr/AhnCPB16} infused factual knowledge into an RNN-based language model to generate Wikipedia-style summary paragraphs of film actors. Similar to ours, their approach also uses a copy mechanism to copy specific words from the input facts to the description. However, these approaches are not directly compatible with our problem setting, which focuses on generating synoptic, rather than detailed multi-sentence descriptions. Additionally, in contrast to our setting, which requires dynamically considering a wide variety of domains and entity types, the previous studies consider just human biographies as a single domain. Our experiments show that our method substantially outperforms the approach by Ahn et al.\ on our task.

The WebNLG Challenge \cite{Gardent-EtAl:2017:INLG2017} is another task aiming at generating text from RDF triples. However, it differs quite substantially from the task we study in this paper, as it demands a textual verbalization of every single triple. Our task, in contrast, requires synthesizing a short synoptic description by precisely selecting the most relevant and distinctive facts from the set of all available facts about the entity.

Bhowmik and de Melo (2018) \cite{BhowmikdeMelo2018} introduced a dynamic memory-based neural architecture that generates 
type descriptions for Wikidata items. Although the study has similar aims as the present one, the approach proposed there mainly exploits the statistical co-occurrence of words in the target descriptions of the training data. In our experiments, we show that the dynamic memory method is unable to be precise in an open-domain setting, which may require invoking named entities such as \emph{Ehime} for the example in Figure~\ref{fig:fogo_gogo}.

\subsection{Referring Expression Generation} Referring Expression Generation (REG) is a subtask of Natural Language Generation (NLG) that focuses on the creation of noun phrases that identify specific entities. The task comprises two steps. The \emph{content selection} subtask determines which set of properties distinguish the target entity, and the \emph{linguistic realization} part defines how these properties are translated into natural language. There is a long history of research on generating referring expressions. In one of the recent approaches, Kutlak et al.\ \cite{kutlak} convert property--value pairs to text using a hand-crafted mapping scheme. However, their method requires specific templates for each domain.  Applying template-based methods to open-domain knowledge bases is extremely challenging, as this would require too many different templates for different types of entities. Recently, Ferreira et al.\ \cite{Ferreira2018} proposed an end-to-end neural approach to REG called NeuralREG. They used a delexicalized WebNLG corpus for the training and evaluation of their model. NeuralREG generates a delexicalized template of the referring expression.

\subsection{Neural Text Summarization} Generating descriptions for entities is related to the task of text summarization in that the salience of information needs to be assessed \cite{YangDeMelo2017HiText}. Similar to abstractive summarization, our task requires the generation of words not seen in the inputs. At the same time, in a similar vein to extractive summarization, our task also requires a selection of words from input facts for copying to the output sequence. The surge of sequence-to-sequence language modeling via LSTMs naturally extends to the task of abstractive summarization by training a model to accept a longer sequence as input and learning to generate a shorter compressed sequence as a summary. 
To this end, Rush et al.\ \cite{rush2015neural} employed this idea to generate a short headline from the first sentence of a text. Recently, Liu et al.\ \cite{LiuEtAl2018GeneratingWikipedia} presented a model that generates an entire Wikipedia article via a neural decoder component that performs abstractive summarization of multiple source documents. 
Our work differs from such previous work in that we do not consider a text sequence as input. Rather, our inputs are a series of property--value pairs deemed as facts.

\subsection{Pointer Networks and Copy Mechanisms}
In order to learn how to solve combinatorial optimization problems that involve output dictionaries of varying sizes, such as the traveling salesman problem, Vinyals et al.\ \cite{NIPS2015:Vinyals} proposed a deep neural architecture known as \emph{Pointer Networks}. Such networks rely on an attention mechanism to repeatedly select elements from the input as output. 
Subsequent works \cite{ACL2016:Gu,ACL2016:Gulcehre,DBLP:journals/corr/MerityXBS16} incorporated this idea into hybrid text-based sequence-to-sequence models that occasionally select words from the input and otherwise generate words from a regular output dictionary. This addresses how to cope with rare and unknown words in the desired output sequence, which constitutes one of the key challenges in deploying neural text generation in practice.
Since it is not feasible to directly learn from existing examples how to generate all possible vocabulary words that might be needed in the output, oftentimes, it is easier to learn to directly select suitable words from the given input.  See et al.\ \cite{DBLP:conf/acl/SeeLM17} investigated the use of such architectures for the task of abstractive text summarization, so as to better cope with long input texts. As mentioned, Ahn et al.~\cite{DBLP:journals/corr/AhnCPB16} rely on a similar copy mechanism to transform facts into summaries. 

While our model adopts similar principles, there is a significant difference in our copy mechanism. In the above-mentioned existing works, the pointing and copying is a one-step process, in which a token within a context window is chosen based on simple attention weights. In our model, the pointing and copying is a two-step process. The model first needs to identify a pertinent fact that is salient enough to merit consideration. Then, within the chosen fact, it selects a suitable word for copying. The context window also varies depending on the number of facts and the number of words within a fact.
In our experiments, we show that our approach greatly outperforms the model with copy mechanism proposed by Ahn et al.

\section{Conclusion}
Synoptic descriptions of entities provide end-users a near-instan\-taneous understanding of the identity of an entity. Generating such descriptions requires the selection of the most discernible facts  and composing these facts into a coherent sequence. In this paper, we introduced a fact-to-sequence encoder--decoder model with a custom copy mechanism. Through an extensive evaluation, we have shown that our method consistently outperforms several competitive baselines on two benchmark datasets. Additional analysis also shows that our model can precisely choose facts and copy factual words effectively. The success of this novel architecture for generating concise descriptions suggests that it could be adapted for additional applications that may benefit from a repository of structured facts, e.g. dialogue systems and knowledge graph-driven question answering.

\begin{acks}
Gerard de Melo's research is in part supported by the Defense Advanced Research Projects Agency (DARPA) and the Army Research Office (ARO) under Contract No.\ W911NF-17-C-0098. Any opinions, findings and conclusions, or recommendations expressed in this material are those of the authors and do not necessarily reflect the views of DARPA and the ARO.
\end{acks}